\def\year{2020}\relax
\documentclass[letterpaper]{article} 
\usepackage{aaai20}  
\usepackage{times}  
\usepackage{helvet} 
\usepackage{courier}  
\usepackage[hyphens]{url}  
\usepackage{graphicx} 
\usepackage{graphicx}  
\usepackage{xcolor}
\usepackage[linesnumbered,vlined,ruled]{algorithm2e}
\SetKwFunction{AvgToFroDist}{Average-To-Fro-Distance}
\SetKwFunction{FastMapD}{FastMap-D}
\SetKwProg{function}{Function}{}{}
\SetKwProg{sub}{\(\bullet\)}{}{}
\usepackage{subcaption}
\usepackage{comment}
\usepackage{amssymb}
\usepackage{amsmath}
\usepackage{amsmath}
\DeclareMathOperator*{\argmax}{arg\,max}
\makeatletter
\newcommand*{\centerfloat}{%
  \parindent \z@
  \leftskip \z@ \@plus 1fil \@minus \textwidth
  \rightskip\leftskip
  \parfillskip \z@skip}
\makeatother

\nocopyright
\title{Integrating Planning, Execution and Monitoring in the presence of Open World Novelties: Case Study of an Open World Monopoly Solver}

\author{Sriram Gopalakrishnan \thanks{indicates equal contribution} \footnotemark[2],  Utkarsh Soni \footnotemark[1] \footnotemark[2],  Tung Thai \footnotemark[3], Panagiotis Lymperopoulos \footnotemark[3], \\ {\bf \Large  Matthias Scheutz \footnotemark[3], Subbarao Kambhampati \footnotemark[2]}
\\
\footnotemark[2] Arizona State University, \\
\{\texttt{sgopal28, usoni1, rao}\}\texttt{@asu.edu} \\
\footnotemark[3] Tufts University}

\begin{document}

\maketitle

\begin{abstract}
    
The game of monopoly is an adversarial multi-agent domain where there is no fixed goal other than to be the last player solvent There are useful subgoals like monopolizing sets of properties, and developing them. There is also a lot of uncertainty from dice rolls, card-draws, and adversaries' strategies. This unpredictability is made worse when unknown novelties are added during gameplay. Given these challenges, Monopoly was one of the test beds chosen for the DARPA-SAILON program which aims to create agents that can detect and accommodate novelties. To handle the game complexities, we developed an agent that eschews complete plans, and adapts it's policy online as the game evolves.  In the most recent independent evaluation in the SAILON program, our agent was the best performing agent on most measures. We herein present our approach and results. 

\end{abstract}






\section{Introduction}

AI agents are often trained and evaluated in closed settings where the dynamics are fixed. They have shown spectacular performance in such settings; this is notably seen in game-playing agents such as in Chess and Go \cite{alphago}. However, we seldom consider how well these agents would act when novelties or changes are injected into the environment, i.e. an open-world setting. This would require execution monitoring to know what parts of the model have changed, and adapting to it as necessary. Developing agents to handle an open-world setting is necessary if we want to bring robust AI-agents into the real world. 

With this in mind, DARPA (Defense Advanced Research Projects Agency) started a research program on Science of Artificial Intelligence and Learning for Open-world Novelty (SAIL-ON). The agents developed as part of this program are developed with the objective of handling novelties in the environment. One of the test-beds chosen for the SAIL-ON program is the game of Monopoly.

Monopoly is a board game about real-estate development with upto 4 adversaries. The objective of the game is to be the last solvent player. This is done through buying and developing properties, so as to charge the other players rent and fees. The game dynamics are first affected by dice rolls which determines how each player moves around the board. The game dynamics are also affected by drawing of ``chance" and ``community cards" ( elements of luck), as well as by the combinations of the actions (strategies) of adversaries. If one adds in {\em novelties}, such as changes to the board layout, rent or bank rates, then the game becomes more unpredictable and hard to pre-train for.

Due to the aforementioned challenges, a plan of action can fall apart in a single round. In such an uncertain environment, we take the cautious and simple approach of state-evaluation after a single-step, where the strength of the approach comes from cautiously approximating the future value of the state after an action. This relatively simple approach outperformed other approaches as evaluated by an independent performer in the DARPA SAIL-ON program on the Monopoly test-bed. 

In this paper, we first present and frame Monopoly as a challenging test-bed for interleaving online planning and execution, especially when novelties are injected (open-world setting). Then we discuss our agent for Monopoly, and also present the results of the evaluation made by an independent third-party evaluator; the evaluation compared our method against other teams in the DARPA SAIL-ON program. We propose our agent methodology as a strong baseline for future research on open-world robustness, and agents in Monopoly.

\section{Monopoly Game And Simulator}

Monopoly is a multi-player adversarial board game  with upto 4 players (traditionally), where the objective is to be the last player solvent after having bankrupted the others. This is done through buying and developing properties so as to charge higher rent when the other players land on your properties. Players move across the board based on dice rolls, and can buy properties owned by the bank. If one lands on a property owned by another player, rent is charged. Rent on a property can be increased by owning all properties of a set (categorized by color); this is called having a \emph{monopoly} over that set. The rent can be further increased by building houses and hotels on a monopolized set. Any policy or plan of action needs to be adapt to the changes of fortune with dice rolls, and the decisions of other players, which makes it challenging as a domain for integrated planning and execution.
The full set of rules for the game simulator (test bed) and all the nuances can be found in \cite{HybridDeepRL_Monopoly} which also contains a link to the game simulator code. This simulator was developed by an independent evaluator for the SAILON program. In the game simulator, one can also inject novelties on top of the standard game to study how the agent adapts to these modifications. Novelties were part of the evaluations of the agents developed for the SAIL-ON program, and will be discussed more in subsequent sections



\subsection{Game Novelty}

For the SAIL-ON program evaluation, agents were tested with one novelty injected per trial, where each trial is 100 games of Monopoly. The novelty could be added in any one of the 100 games and persists for the remaining games. The novelty could be changing the number of properties in a set required for Monopoly, the rent of a property after building a hotel on it, the order of properties on the board and such. The set of possible novelties is \emph{not} shared with us by the evaluation/test team, and so it is left to us to make the agent as robust  and adaptable to novelties as possible.

\section{Agent Design}


We developed our agent such that it's policy is controlled by a state-value function. The value of a state is primarily determined by the expected short and long term reward obtained from that state. Before we discuss the details of how these rewards are calculated, we first present the motivation for our design. Typically, approaches that use a value function for game-playing agents -- like MCTS \cite{MCTS_survey}-- either simulate trajectories to the end and backpropagate the terminal state value to compute the starting state's value, or they use a limited lookahead with an evaluation function that captures the value of the rest of the trajectory. We use the simplest form of the limited-lookahead approach where we just lookahead by 1-step and then evaluate the next state by approximating the expected short and long term returns that would result by taking the action.

Our reason for planning with a 1-step lookahead was that in the game of Monopoly, a single roll of the die, or a chance card, or an adversary's decision could change the entire value of a state. So to compute the value of a state accurately with simulated actions, requires considering a very large set of branches from an extremely wide and deep tree that includes many possible combinations of dice rolls, combination of player decisions, auction bids, and more. It should be noted that each turn of a player also includes what are called out-of-turn moves by other players, which further increases the branching factor of the search tree; please refer to the monopoly rules in \cite{HybridDeepRL_Monopoly} for more detailed information. 

If one had a very accurate mental model of adversaries, the possible branches of the search tree might become more manageable. Additionally, pre-training a large neural network for state evaluation--as was done with alpha-go \cite{alphago}-- is not viable since our agent would have to handle novelties or modifications to the game (the space of which we did not know).
Lastly, the evaluators impose a max time limit of 3 hours per full-game, so simulating enough MCTS rollouts for each action did not seem feasible.

Thus, in this work, our focus on intelligently evaluating a state after a single action by considering short and long term consequences; rather than requiring an accurate and complete model to rollout and evaluate each state, we consider long term consequences with simplifying assumptions (will be discussed). The state value includes the current monetary value of possessions,  potential short and long term gains, as well as the future benefit of monopolized properties. Importantly, the evaluation function is largely parameterized with game attributes (that can change) and has few tuned constants; this helps make it robust to game variations. We will first go over the evaluation function. We will then provide some examples of the state attributes that are tracked and updated in $\mathcal{V}(s)$ to accommodate for novelty. 


\subsection{State Evaluation Function}



The value of a state should consider the current (monetary) value of owned properties as well as the potential for future earning as possible future rewards. Thus, the evaluation function we propose is a linear combination of four terms i.e. $\mathcal{V}(s) = \mathcal{M}_\text{assets} +\mathcal{R}_{\text{s}} + \mathcal{R}_{\text{l}} + \mathcal{M}_{\text{monopoly}}
$. Each of these terms is described below:


\textbf{$\mathcal{M}_\text{assets}$:} Property value of all the agent's assets that are not currently mortgaged. Each property can be mortgaged with the bank for cash. We can buy back the property from the bank for the mortgaged amount plus interest on the mortgage.

\textbf{$\mathcal{R}_{\text{s}}$}: short term expected gain in funds computed as the difference between expected rent the agent will get for the properties that it owns in state $s$ and the expected rent it would owe to other players based on current ownership of properties over the next $k$ turns. The expectations are computed over the probabilities of each player landing in a particular position in the next $k$ turns. This is akin to a rollout with the strong relaxation (assumption) that no more properties will be bought or developed. To be specific, let $\mathbb{G}$ be the set of all agents, $g_{1}$ be our agent, $\mathcal{P}(g)$ denote the properties owned by agent $g$, $r(p)$ denote the rent of property $p$, and $Pr(g, p, k)$ denote the probability that an agent $g$ will land on a property $p$ in the $k^{\text{th}} $ turn from state $s$, then $\mathcal{R}_{s}$ is computed as $\mathcal{R}_{s} = \sum_{k} \sum_{g \in \mathbb{G} - g_{1}} \Big[\sum_{p \in \mathcal{P}_{(g_{1})}} Pr(g, p, k) * r(p) - \sum_{p \in \mathcal{P}_{(g)}} Pr(g_{1}, p, k) * r(p)$\Big].

\textbf{$\mathcal{R}_{\text{l}}$}: the expected long term change in funds. The computation for this term is similar to $\mathcal{R}_{\text{s}}$, except that the probability of an agent landing on a property is assumed to be uniform over all properties. Note that the long term gain is calculated for $k$ full loops/passes around the board (not turns). The value of $k$ for both $\mathcal{R}_{\text{s}}$ and $\mathcal{R}_{\text{l}}$ was taken as $5$.

\textbf{$\mathcal{M}_{\text{monopoly}}$}: A monopoly gain term is computed to incorporate the monetary benefit our agent would get for monopolizing and improving all properties of the same color. The purpose of this term is to drive our agent towards taking actions that would let it gain a monopoly on a color and subsequently perform maximal improvements on its properties. To compute $\mathcal{M}_{\text{monopoly}}$, we start by calculating the expected funds, $\mathcal{F}$, our agent would have after going around the board (full loop) $k$ times ( $k=5$ in our implementation) from its current position as 
    $\mathcal{F} = \text{cash possessed} + k * \text{go\_increment} + \mathcal{R}_{\text{l}}$
    where the last term $\mathcal{R}_{\text{l}}$ is also computed for $k$ loops around the board. Now let $\mathbb{C}(g_{1})$ be the set of all colors such that our agent owns at least one property of that color. Then for each $c \in \mathbb{C}(g_{1})$, we compute the combined potential rent for that color ($\mathcal{R}_{c}$) that our agent will get from all the properties of that color if it spends all of $\mathcal{F}$ in buying all the properties of color $c$ followed by improving each of the properties as much as possible with the remaining amount from $\mathcal{F}$. This potential rent value is then scaled down based on how many properties the agent actually possess (currently) for that color. For example, if we own 1 out of 3 blue properties, then the potential value from that color should be much less than if we own 2 out of 3 red properties. The scaled potential value $\mathcal{R}^{s}_{c}$ is computed as
    $\mathcal{R}^{s}_{c} = \mathcal{R}_{c} / 2^{\mathcal{P}_{(c)} - \mathcal{P}_{(g_{1})}}$
    where $\mathcal{P}_{(c)}$ is the total number of properties of color $c$. We use an exponential function in the denominator to value color sets that are closer to completion significantly more than others. Since the set size can change as part of game novelties, we think this is prudent. Finally, the monopoly component of the state evaluation, $\mathcal{M}_{\text{monopoly}}$, is simply computed as the maximum  $\mathcal{R}^{s}_{c}$ over all the $c \in \mathbb{C}(g_{1})$. What this monopoly term does for the agent is to allow it to eschew buying new or bidding for properties if that amount can be used to complete and develop a monopoly.

\subsection{Avoiding Bankruptcy}
Another complication for the agent, is that it must try to avoid bankruptcy in the face of a lot of stochasticity from the game. So even if the expected value of a policy is high, if it risks bankruptcy then a lesser-value policy that minimizes the risk of bankruptcy might be preferred. Concretely, at any state $s$, the agent considers if each possible move from the set of possible moves $m \in \mathbb{M}$ with cost $\mathcal{C}(m)$ satisfies the following conditions: 

\textit{Condition 1:} $cash_{\text{current}} + \mathcal{R}_{\text{next}}  - \mathcal{C}(m) \geq cash_{\text{min}}$ where $cash_{\text{current}}$ is the current amount of money our agent possess, $\mathcal{R}_{\text{next}}$ is the expected change in cash due to rent after one round, $cash_{\text{min}}$ denotes the absolute minimum amount needed to protect against bankruptcy. This covers misfortunes from the Chance and Community chest cards that the agent might draw.
    
\textit{Condition 2:} $cash_{\text{current}} + \mathcal{R}_{\text{owed}} + worth_{\text{scaled}} - \mathcal{C}(m) - \mathcal{R}_{\text{worst}} > 0$ where $\mathcal{R}_{\text{owed}}$ is the expected income from charging rent that our agent will get in the next round, $worth_{\text{scaled}}$ is some mortgage value of all properties our agent owns, and $\mathcal{R}_{\text{worst}}$ is the maximum possible rent our agent could be charged in the next round. This protects against bankruptcy from landing on an adversary's property.
Both the above conditions were used to prevent the agent from aggressively spending its cash and going bankrupt. Once we have pruned the moves in $\mathbb{M}$, our agent simply chooses the move such that $m = \argmax_m V(s'_{m})$ where $s'_{m}$ is the next state after simulating the move $m$.

\subsection{Novelty Detection and adaptation}

To perform well in the SAIL-ON evaluation, our agent needs to detect the novelties introduced and adapt state evaluation accordingly. The novelties that the agent was tested on were hidden. To adapt, we maintain knowledge of the expected values for game-board attributes like property rent, dice outcome likelihood, etc. The evaluation function is parameterized with such attributes and is updated once a change is detected. 
Some of these values are provided directly as the state information and thus we keep track of the current values of these attributes by observing the state. For other attributes, the agent needs to observe the outcome of certain actions (like selling a property) to infer how the relevant attributes changed. In attribute value changes, we also detect novelties related to dice. This includes addition/deletion of a die, additional sides added to the dice, and the distribution of rolling any number on each die. The first two are inferred by observing the dice rolls in the game. For the last one, we model the distribution of rolling any number as Dirichlet distribution and use MAP estimates to update this distribution. This updated dice distribution is then used to compute the probability function $Pr$ , as used to compute $\mathcal{R}_s$. 

\section{Evaluation and Results}

\begin{table}[!ht]
\scriptsize
\centering
\begin{tabular}{|l|l|l|l|l|}
\hline
Novelty type & \begin{tabular}[c]{@{}l@{}}NDA(\%)\end{tabular} & \begin{tabular}[c]{@{}l@{}}Best\\ competitor \\ NDA(\%)\end{tabular} & Win-rate (\%) & \begin{tabular}[c]{@{}l@{}}Best\\competitor\\ win \\rate (\%)\end{tabular} \\ \hline
None(PNWP) & \multicolumn{1}{c|}{-} & \multicolumn{1}{c|}{-} & 76.48 & 63.61 \\ \hline
CN-easy & 40 & 20 & 71.1295 & 58.448 \\ \hline
CN-medium & 46.67 & 20 & 64.8895 & 61.867 \\ \hline
CN-hard & 48.00 & 24 & 28.899 & 43.173 \\ \hline
AN-easy & 90.32 & 80 & 70.473 & 62.335 \\ \hline
AN-medium & 90 & 90 & 94.432 & 68.5425 \\ \hline
AN-hard & 75 & 20 & 61.503 & 46.1825 \\ \hline
RN-easy & 52 & 3.33 & 74.9905 & 58.448 \\ \hline
RN-medium & 32 & 13.33 & 78.0975 & 50.8625 \\ \hline
RN-hard & 32 & 20 & 81.5815 & 67.6585 \\ \hline
\end{tabular}
\caption{Novelty detection and reaction performance in the DARPA SAILON program}
\label{tab:results}
\end{table}

Our agent was evaluated against other teams in the SAIL-ON program by an independent evaluator separately funded by DARPA. Evaluation consists of multiple trails, where each trial consists of $100$ games of monopoly against 3 baseline agents. The baseline adversarial agents were programmed by the evaluation team to serve as the competition baseline. The behavior of the baseline agent is described in \cite{HybridDeepRL_Monopoly} as the "simple baseline agent" in that work. During each trial, a novelty is injected during one of the 100 games, and kept for the remaining games.

\subsection{Measures of performance} 
The following metrics help compare agent performance:
(1) Pre-Novelty Win-Percentage (PNWP): The ratio of games won before any added novelty. (2) Novelty Detection Accuracy (NDA): This is the percentage of trials in which the novelty was correctly detected, and without a false positive before the novelty was added. (3) Novelty Reaction Performance (NRP): To compute this, the win ratio of our agent after the novelty was added is divided by the win ratio of the baseline agent before the novelty was added. 
These measures were not defined by us, but by the evaluation group, and directed by discussions in the SAIL-ON program.
For every measure, our agent was evaluated with different classes of novelties, these are: Class Novelties (CN) such as new classes of objects like property classes, or new classes of actions; Attribute Novelties (AN) such as changes in the mortgage rate, rent costs; and Representation Novelties (RN) such as changes to the position of properties, and the color sets to which they belonged. Within each type of novelty, the evaluators further classified them into easy, medium and hard. As mentioned, we do not have more details about the specific type and distribution of novelties that our agent was evaluated on, as this information is currently hidden from us to evaluate agent adaptation better.
\subsection{Results}

We report our performance in Table \ref{tab:results} where we provide our performance and the performance of the best competitor for NDA and Win percentages for the different novelty settings. With respect to the win ratio of our agent, our pre-novelty win ratio (PNWR) was 76.48\%, i.e the 3 other baseline agents combined only won less than a quarter of the games when there was no novelty injected. This win rate represents the efficacy of our agent design/playing algorithm for the standard Monopoly game. In comparison the win rate for the next best team was 63.61\%. Our agent performs better before novelty was added to the game, and also in most settings after novelty was added to the game. The only setting in which our agent did not get the best result was for "NRP-CN-hard" (Novelty Reaction performance for hard class- novelties). This reflects our agent's ability to accurately capture both the short and long terms effects of actions, as well as, how well it can adjust for novelties in the game while making decisions by modifying the evaluation function during the gameplay. The evaluators also ran a special test to see how well our agent performs against another instance of our agent, and a baseline agent. The two instances of our agent won 40.75 and 39.43\% of the games on average (over many trials). 

\section{Related Work}

There are connections between replanning systems and handling open-world novelty. Replanning systems are aimed at dealing with unanticipated changes in the dynamics, but traditionally, replanning systems don’t automatically characterize the novelty or change their domain model \cite{FFreplan}. In \cite{cushing2005replanning}, the authors discuss how to update the planning problem to handle unexpected changes with a language for failure representation, but not how to automatically characterize a novelty and update the model. In our methodology we do online model-update and replanning by incorporating the novelties into our state-evaluation function.
 
With respect to agents for Monopoly, there have only been a few notable attempts;  \cite{HybridDeepRL_Monopoly} recently proposed a Reinforcement Learning (RL) approach where they train a Deep Q-Network agent \cite{mnih2015human} to play Monopoly. To accelerate learning, they employ a $\epsilon$-greedy approach during training where instead of executing a random action for exploration, the agent imitates the policy of a rule-based agent which was manually designed to follow known successful strategies to winning Monopoly. Other RL-based approaches include \cite{bailis2014learning} and \cite{arun2019monopoly} where the Q-function is again approximated by a neural network. To train the agent, the former approach uses the $Q(\lambda)$-learning technique \cite{peng1994incremental}, whereas the latter uses experience replay \cite{mnih2015human}. To restrict the action space of the agent, all the mentioned techniques only select the action type, and the parameters are chosen by fixed rules; for example the \textit{sell\_property} action in \cite{HybridDeepRL_Monopoly} would sell possesions in a fixed order, a hotel, a house or a property depending on availability.) In our approach, we only constrain the trade-related actions to rule based behavior. Further, we argue that our approach is more suited to developing an agent that is robust to open-world pertubations; the aforementioned RL-based approaches would require retraining the agent once any novelty is encountered, even some simple parameter changes such as rent values.

Another approach for playing Monopoly was presented in \cite{sammul2018learning} which uses MCTS \cite{MCTS_survey} for its decisions. To make this feasible, the author makes significiant game simplifications to reduce the action space and game-tree size. For example no out-of-turn actions are allowed, which is a significant deviation from the game. This pruning, coupled with the agent having access to the best (by win rate) adversary model it will play against, is what helped the MCTS method achieve a win rate of 63\%. We, on the other hand, handle the chaotic nature of the game by evaluating the state features intelligently, and not rolling-out and relying on the availability of an accurate adversary model.

\section{Conclusions and Future Work}
We present the game of Monopoly as a challenging test bed for evaluating open-world robustness, and integrating online planning and execution. We propose our agent methodology of using a flexible state evaluation function as a strong approach to handling novelties in the environment as demonstrated through an independent evaluation in the game of Monopoly. There are plenty of interesting avenues for further research: these include including learning and updating adversary mental models, and analyzing the cost/benefit tradeoff in varying the lookahead depth in the game tree especially if the game and adversary models change over time. 

\section{Acknowledgement}
This research is supported by the DARPA SAIL-ON grant W911NF-19-2-0006.


\bibliography{references}
\bibliographystyle{aaai}
\end{document}